\documentclass[letterpaper, 10 pt, journal, twoside]{IEEEtran}

\usepackage{graphics}           
\usepackage{times}              
\usepackage{amsmath}            
\usepackage{amssymb}            
\usepackage{graphicx}
\usepackage{algorithm}
\usepackage[noend]{algpseudocode}
\usepackage{booktabs}
\usepackage[dvipsnames, svgnames, x11names]{xcolor}
\usepackage{color}
\usepackage{subfigure}
\usepackage{multirow}
\usepackage{setspace}
\definecolor{instructioncolor}{rgb}{.5,.5,.5}
\usepackage{threeparttable}
\usepackage{stmaryrd}
\usepackage{makecell}
\usepackage{cite}
\usepackage{bbding}
\usepackage{colortbl}
\makeatletter
\let\NAT@parse\undefined
\makeatother
\usepackage[colorlinks=true,
            linkcolor=black, 
            anchorcolor=black, 
            citecolor=black,
            urlcolor=black
            ]{hyperref}

\makeatletter
\renewcommand{\maketag@@@}[1]{\hbox{\m@th\normalsize\normalfont#1}}%
\makeatother

\usepackage[font=small]{caption}

\def\secref#1{Sec.~\ref{#1}}
\def\figref#1{Fig.~\ref{#1}}
\def\tabref#1{Tab.~\ref{#1}}
\def\eqref#1{Eq.~(\ref{#1})}

\makeatletter

\newcommand{\Rmnum}[1]{\expandafter\@slowromancap\romannumeral #1@}
\makeatother

\makeatletter
\usepackage{xspace}
\DeclareRobustCommand\onedot{\futurelet\@let@token\@onedot}
\def\@onedot{\ifx\@let@token.\else.\null\fi\xspace}

\def\etal{{et al}\onedot}
\makeatother

\def\etalcite#1{\etal~\cite{#1}}

\usepackage{array}
\newcolumntype{L}[1]{>{\raggedright\let\newline\\\arraybackslash\hspace{0pt}}m{#1}}
\newcolumntype{C}[1]{>{\centering\let\newline\\\arraybackslash\hspace{0pt}}m{#1}}
\newcolumntype{R}[1]{>{\raggedleft\let\newline\\\arraybackslash\hspace{0pt}}m{#1}}

\newcommand{\RR}{\mathbb{R}}

\renewcommand{\b}[1]{\mbox{\boldmath$#1$}}

\newcommand{\tr}[0]{\sf T}              

\usepackage{url}

\newcommand\difffirst[1]{\textcolor{black}{#1}}

\title{SGLC: Semantic Graph-Guided Coarse-Fine-Refine Full Loop Closing for LiDAR SLAM}

\author{Neng Wang, Xieyuanli Chen$^\dag$, Chenghao Shi, Zhiqiang Zheng, Hongshan Yu, Huimin Lu$^\dag$
  \thanks{Manuscript received: July 6, 2024; Revised: August 23, 2024; Accepted: October 22, 2024. This paper was recommended for publication by Editor \mbox{C. Javier} upon evaluation of the Associate Editor and Reviewers’ comments.}%
  \thanks{This work was supported in part by the National Science Foundation of China under Grant 62403478, Young Elite Scientists Sponsorship Program by CAST (No. 2023QNRC001), as well as Major Project of Natural Science Foundation of Hunan Province under Grant 2021JC0004.}
  \thanks{N. Wang, X. Chen, C. Shi, Z. Zheng and H. Lu are with the College of Intelligence Science and Technology, National University of Defense Technology. H. Yu is with the Hunan University.  $^\dag$joint corresponding author: xieyuanli.chen@nudt.edu.cn, lhmnew@nudt.edu.cn.}
  \thanks{Digital Object Identifier (DOI): see top of this page}
  
}

\begin{document}
\maketitle
\markboth{IEEE ROBOTICS AND AUTOMATION LETTERS. PREPRINT VERSION. ACCEPTED OCTOBER, 2024}%
{Wang \MakeLowercase{\textit{et al.}}: SGLC: Semantic Graph-Guided Coarse-Fine-Refine Full Loop Closing for LiDAR SLAM}


\begin{abstract}
    Loop closing is a crucial component in SLAM that helps eliminate accumulated errors through two main steps: loop detection and loop pose correction. The first step determines whether loop closing should be performed, while the second estimates the 6-DoF pose to correct odometry drift. Current methods mostly focus on developing robust descriptors for loop closure detection, often neglecting loop pose estimation.
    A few methods that do include pose estimation either suffer from low accuracy or incur high computational costs.
    To tackle this problem, we introduce SGLC, a real-time semantic graph-guided full loop closing method, with robust loop closure detection and 6-DoF pose estimation capabilities.
    SGLC takes into account the distinct characteristics of foreground and background points.
    For foreground instances, it builds a semantic graph that not only abstracts point cloud representation for fast descriptor generation and matching but also guides the subsequent loop verification and initial pose estimation.
    Background points, meanwhile, are exploited to provide more geometric features for scan-wise descriptor construction and stable planar information for further pose refinement.
    Loop pose estimation employs a \mbox{coarse-fine-refine} registration scheme that considers the alignment of both instance points and background points, offering high efficiency and accuracy.
    \difffirst{Extensive experiments on multiple publicly available datasets demonstrate its superiority over state-of-the-art methods.}
    Additionally, we integrate SGLC into a SLAM system, eliminating accumulated errors and improving overall SLAM performance.
    The implementation of SGLC will be released at \url{https://github.com/nubot-nudt/SGLC}.
\end{abstract}


\section{Introduction}
\label{sec:intro}
Loop closing plays a crucial role in simultaneous localization and mapping (SLAM) systems for correcting odometry drifts and building consistent maps, especially in single-sensor SLAM without GPS information.
LiDAR-based loop closing typically requires feature extraction from LiDAR scans to generate discriminative descriptors. This task is especially challenging due to the extensive and sparse nature of point clouds in outdoor operations for autonomous robots.

\begin{figure}[ht]
	\centering
	\includegraphics[width=1\linewidth]{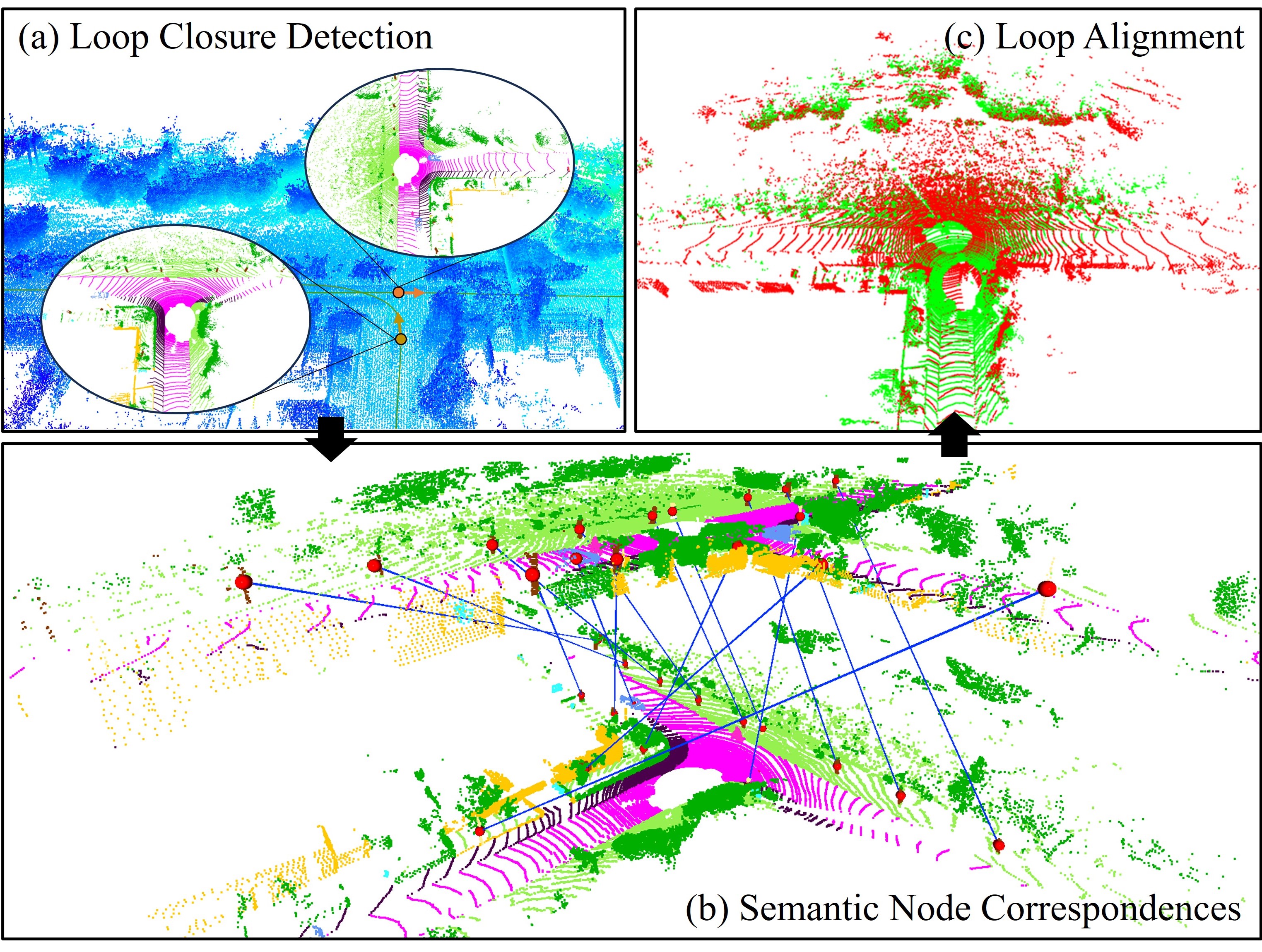}
	\caption{Visualization of loop closing using our method. 
    (a) Loop closure detection, it shows a reverse loop on the KITTI 08 sequence found by our approach even with significant changes in the position and orientation. 
    (b) Semantic node correspondences for geometric verification and initial loop poses estimation. The blue lines  indicate the node correspondences and red spheres represent the estimated instance center.
    (c) Final alignment for loop correction.
	}
	\label{fig:motivation}
	\vspace{-0.2cm}
\end{figure}

While many existing methods~\cite{He2016iros,Uy2018cvpr,Kong2020iros,Wang2020icraisc,ma2022ral,Luo2023iccv} focus on loop closure detection (LCD), few can estimate closed-loop pose in six degrees of freedom (6-DoF). Some methods incorporate pose estimation with 1-DoF~\cite{kim2022tro,wang2020irosiris,chen2021auro} or 3-DoF~\cite{li2021iros,Jiang2023icra}, which may not be sufficient for 6-DoF SLAM systems. \difffirst{Existing} full  6-DoF pose estimation methods, however, are either very time-consuming~\cite{Cattaneo2022tro} or have relatively lower accuracy~\cite{Cui2022ralO}.

To tackle these issues, we propose SGLC, an efficient semantic graph-guided full loop closing framework that offers both robust LCD and accurate 6-DoF loop pose estimation.
Different from existing semantic graph-based methods~\cite{Kong2020iros,Qiao2023iros} that indiscriminately use both foreground instances and background points, or those overlook the geometrically rich background point clouds~\cite{zhuy2020iros}, our method leverages the distinct properties of both foreground and background elements efficiently. SGLC builds semantic graphs based on foreground instances as they can be naturally represented as individual nodes. While for retrieving loop candidates and estimating 6-DoF poses, SGLC exploits both the topological properties of the semantic graph and the geometric features of background points to enhance loop closing accuracy and robustness.

Specifically, SGLC first generates LiDAR scan descriptors to quickly retrieve multiple candidate scans by exploiting the semantics and typologies of the foreground semantic graph and the geometric features of the background. 
To prevent incorrect loops from compromising the SLAM system, we apply geometric verification to eliminate false loop candidates by identifying graph node correspondences between the query and candidate scans. During node matching, we ensure accurate correspondences and facilitate the process through an outlier correspondence pruning method based on their neighbor geometric structure.
Finally, for the verified loop candidate scan, we propose a coarse-fine-refine registration strategy for estimating 6-DoF pose between it and the query scan. This initially estimates a coarse pose by aligning the sparse matched node centers. Then, a fine registration of dense instance points is applied. Additional planar information from the background points further refines the final pose estimation using point-to-plane constraints.
This strategy aligns both foreground instances and background points, with each stage starting from a favorable initial value, ensuring accuracy and efficiency.
\figref{fig:motivation} illustrates how our method accurately detects loop closures and identifies instance node correspondences, even in places with significant direction and position differences, thus finally estimating the loop pose and closing the loop.

In summary, our contributions are fourfold:
(i) We propose a novel semantic graph-guided full loop closing framework, SGLC, that offers robust loop detection and accurate 6-DoF pose estimation.
(ii) We design an effective and efficient outlier pruning method to remove incorrect node correspondences.
(iii) We proposed a coarse-fine-refine registration scheme that improves the accuracy and efficiency of pose estimation.
(iv) We seamlessly incorporate SGLC into a current SLAM framework, reducing the error in odometry estimations.
\difffirst{Extensive evaluation results on KITTI~\cite{Geiger2013ijrr}, KITTI-360~\cite{liao2022tpami}, Ford Campus~\cite{Pandey2011ijrr} and Apollo~\cite{Lu2019cvpr} datasets support these claims.}

\section{Related Work}
\label{sec:related}

\textbf{Handcrafted-based approaches.} 
These methods typically utilize the geometric features within the point cloud to describe it.
In the early stages, Magnusson~\etalcite{Magnusson2009icra} explore the application of the Normal Distributions Transform (NDT) for surface representation to develop histogram descriptors based on surface orientation and smoothness.
The descriptors perform well in different environments and provide the inspiration for subsequent NDT-based methods~\cite {Zaganidis2019iros,Zhou2021icra}.
To quickly extract descriptors, some methods project point clouds into 2D plane~\cite{wang2020irosiris,Wang2020icraisc,li2021iros,kim2022tro} for encoding features. 
Among them, Scan Context (SC) family~\cite{Wang2020icraisc,kim2022tro,li2021iros} is widely used due to its high efficiency and good LCD performance.
It converts the raw point cloud into a polar coordinate bird's eye view (BEV) and encodes the maximum height into the image bins.
This yields rotation-invariant ring key descriptors for retrieval and a more detailed 2D image matrix for calculating similarity.
\difffirst{Based on that, Jiang~\etalcite{Jiang2023icra} employed a similar projection-based approach, but it generates multi-layered BEV images, which can effectively reduce the loss of height information.}
Recently, Cui~\etalcite{Cui2022ralO} propose BoW3D, leveraging Link3D features~\cite{cui2024link3d} to build bag of words and estimate 6-DoF pose.

\textbf{Deep Neural Networks (DNN)-based approaches.}
These methods leverage networks to extract local features and then aggregate them into a global descriptor for retrieval.
The early DNN-based work PointNetVLAD~\cite{Uy2018cvpr} is a combination of PointNet~\cite{qi2017cvpr} for extracting features and NetVLAD~\cite{arandjelovic2016cvpr} for generating the final descriptor.
After this,  a variety of learning-based loop closing methods begin to proliferate.
Liu~\etalcite{Liu2019iccv} proposed an adaptive feature extraction module and a graph-based neighborhood aggregation module for enhancing PointNetVLAD.
In contrast to them, Komorowski~\etalcite{Komorowski2021wacv} build a sparse voxelized point cloud representation and extract features by sparse 3D convolutions.
Chen~\etalcite{chen2021auro} propose OverlapNet, a range image-based LCD method by estimating image overlap.
A subsequent enhanced version, OverlapTransformer~\cite{ma2022ral} is proposed with superior performance and higher efficiency. 
Cattaneo~\etalcite{Cattaneo2022tro} proposed LCDNet, a robust LCD and 6-DoF pose estimation network validated in various environments, but its high computational complexity makes real-time running challenging.
Recently, Luo~\etalcite{Luo2023iccv} proposed a lightweight BEV-based network showing high efficiency and LCD capability.

Some methods also incorporate semantic information to enhance the descriptor distinctiveness.
Kong~\etalcite{Kong2020iros,li2021icrasaloam} and Zhu~\etalcite{zhuy2020iros} attempt to build semantic graphs for loop closing, while Li ~\etalcite{li2021iros} utilize semantics to improve existing descriptors.
\difffirst{Building upon LCDNet~\cite{Cattaneo2022tro}, Arce~\etalcite{Arce2023ral} integrate semantic information into the network only during the training phase, which improves loop closing performance and makes the model versatile.}
We agree that semantics are beneficial for loop closing, as they help differentiate objects in the scene. However, few semantic-based methods have proven both efficient and effective for 6-DoF loop closing.


\begin{figure*}[ht]
	\centering
	\includegraphics[width=1\linewidth]{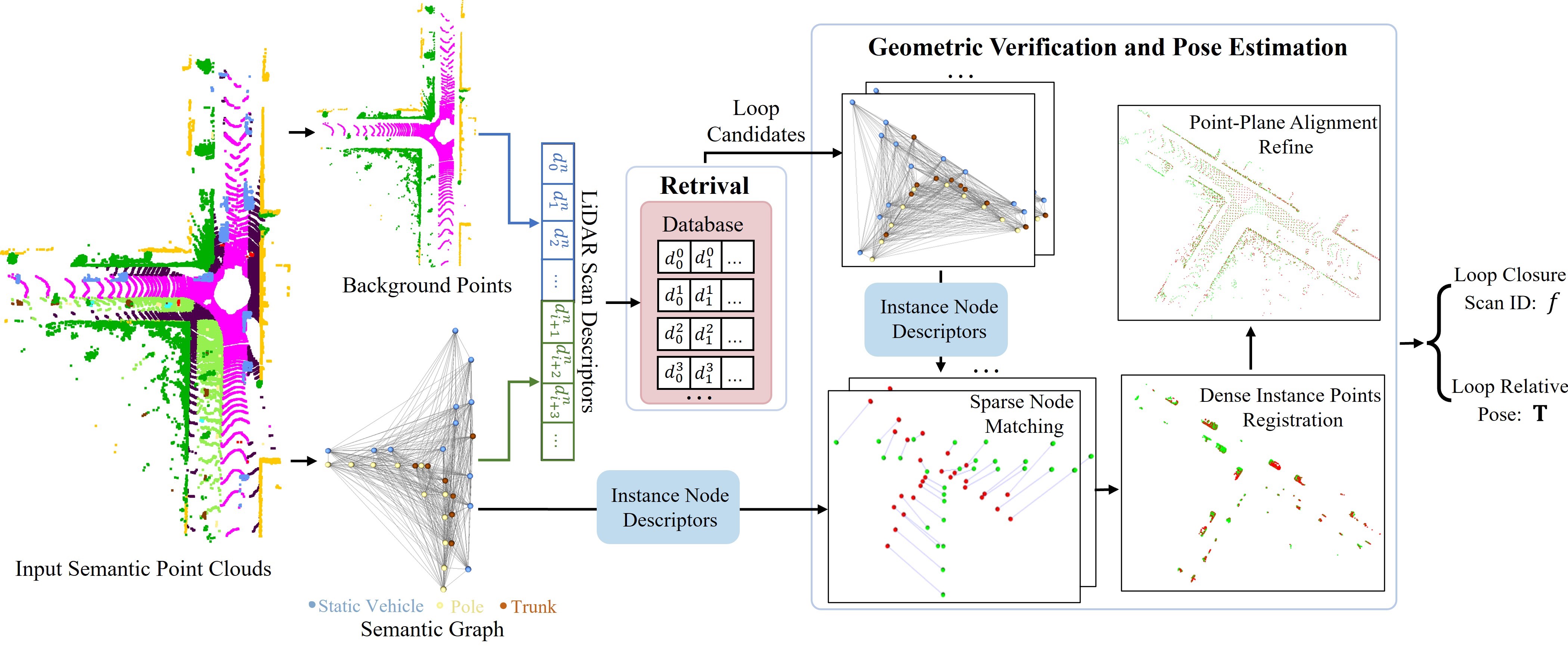}
    \setlength{\abovecaptionskip}{-0.4cm}
	\caption{The framework of SGLC. 
    It first builds a semantic graph for foreground instances and then generates LiDAR scan descriptor considering both the topological properties of the semantic graph and the appearance characteristics of the background.
    The LiDAR scan descriptor is utilized to retrieve loop candidate scans from the database.
    Following this,  geometric verification is performed on each loop candidate to filter out false loop closure, with the key step utilizing the instance node descriptors for robust sparse node matching.
    Finally, a coarse-fine-refine registration scheme is employed to estimate the precise 6-DoF pose.
	}
	\label{fig:pipeline}
	\vspace{-0.4cm}
\end{figure*}

\section{Semantic Graph-Guided Loop Closing}
\label{sec:main}
This section details our semantic graph-guided loop closing approach, dubbed SGLC shown in~\figref{fig:pipeline}, which includes building graphs from the raw point clouds~(\secref{sec:pre}), LiDAR scan descriptor generation~(\secref{sec:global_d}), geometric verification~(\secref{sec:geometric_vgerification})  and pose estimation~(\secref{sec:pose}).

\vspace{-0.1cm}
\subsection{Semantic Graph}
\label{sec:pre}
Semantic graph is a fundamental component of our SGLC, which generates distinctive descriptors and guides the subsequent geometric verification and pose estimation. Given a raw LiDAR scan $\mathcal{S}$, semantic label for each point can be obtained using an existing semantic segmentation method~\cite{wang2024arxiv} with the ability to distinguish between moving and static objects. We then apply clustering~\cite{park2019iros} to such semantic point clouds to identify object instances. 
Subsequently, the bounding box of instances can be estimated by enclosing these clusters.
As foreground instances such as $pole$ and $trunk$ are naturally standalone, they can be easily represented as individual nodes in the graph.
Background point clouds, such as $building$ and $fence$, typically have extensive points and are viewpoint-dependent, making them unstable for representation as instance nodes.
Therefore, we construct semantic graphs only for stable foreground instances, such as $pole, \; trunk, \; lamp, \; static \;vehicle$.
Each node $\mathbf{v}$ comprises instance center position $\mathbf{c} \hspace{-0.2cm}=\hspace{-0.2cm}[x, y, z]^{\tr} $, bounding box size $\mathbf{b} \hspace{-0.05cm}=\hspace{-0.05cm} [l, h, w]$, a semantic label $l_v$, and a node descriptor $\mathbf{f} \in \RR^D$ generated based on the semantic graph typologies. 

Graph edges are established between pairs of nodes if their spatial Euclidean distance is less than $d_{\text{max}}$.
Each edge $\mathbf{e} = (\mathbf{v}_i,\mathbf{v}_j)$ is described by a label $l_e$ determined by the two nodes it connects and the length $d$. The label $l_e$ include categories such as $pole$-$lamp$, $pole$-$trunk$, and so on.
Finally, we obtain the semantic graph $\mathcal{G}$ with a set of nodes $\mathcal{V}$ and a set of edges $\mathcal{E}$.
The graph adjacency matrix $\mathbf{A} \in \RR^{N\times N}$ is created as:
\begin{equation}
\mathbf{A}_{ij} = \begin{cases} 1 & \text{if}(\mathbf{v}_i,\mathbf{v}_j)\in \mathcal{E}, \\
                       0 & \text{otherwise}
         \end{cases}  
\end{equation} 

Based on the semantic graph, we design instance node descriptors for the subsequent robust node matching. 
We first encode the local relationships of nodes by using connected edges.
These edges are categorized and quantified to construct a histogram-based descriptor. \difffirst{Specifically, for a node $\mathbf{v} \in \mathcal{V}$, we 
categorize all connected edges into different intervals based on their labels and lengths, and count the occurrences within each interval with length $d_i$ to generate the descriptor $\mathbf{f}_l$.}

Since $\mathbf{f}_l$ only captures the local typologies within the node neighborhood, it does not contain node global properties, such as their global centrality.
To enhance the distinctiveness of the descriptor for robust node matching, 
we employ eigenvalue decomposition on the graph adjacency matrix $\mathbf{A}$, yielding $\mathbf{A} = \mathbf{Q}\b\Lambda\mathbf{Q}^{\text{T}}$. The eigenvalues in $\text{diag}(\b\Lambda) = {\lambda_1, \ldots, \lambda_n}$ are arranged in descending order.
The $i$-th row of $\mathbf{Q}$ represents the $i$-th instance node embeddings, which capture the node's global properties in the graph~\cite{Nikolentzos2017aaai}. 
Therefore, we use the each row of $\mathbf{Q}$ as part of the instance node descriptor.
Considering that the dimensions of the matrix $\mathbf{Q}$ generated by each graph are different, 
and to unify dimensions of the descriptor, we encode each node using the first $k$ columns of $\mathbf{Q}$, i.e., the eigenvectors corresponding to the largest $k$ eigenvalues, thereby generating a row vector $\mathbf{f}_g \in \RR^k$ for each node.
Note that the signs of eigenvectors are arbitrary, hence we use their absolute values.
Finally, by concatenating $\mathbf{f}_l$, $\mathbf{f}_g$, we obtain each instance node's descriptor $\mathbf{f} \in \RR^D$ in the graph.

\subsection{LiDAR Scan Descriptor for Retrival}
\label{sec:global_d}
Based on our semantic graph, our method effectively detects the loop closure. Theoretically, we could use local node descriptors to determine graph similarity and find loops directly. However, graph matching using nodes for LCD on a scan-scan basis is time-consuming, making it unsuitable for real-time SLAM systems. For fast loop candidate retrieving, we design a novel global descriptor for each LiDAR scan, including the foreground descriptor $\mathbf{F}_{f}$ and background descriptor $\mathbf{F}_{b}$. 

We generate $\mathbf{F}_{f}$ based on our proposed foreground semantic graph, using the graph edges and nodes.
The first part of $\mathbf{F}_{f}$ is similar to the first part of node descriptor $\mathbf{f}_{l}$ as an edge-based histogram descriptor.
Instead of considering only the edges connected to a particular node, for $\mathbf{F}_{f}$, we account for all edges within the entire graph.
These edges are categorized into subintervals in the histogram based on their types and lengths, forming the global edge descriptor.
For the second part, We tally the number of nodes with different labels in the semantic graph to effectively describe the node distribution.

Besides $\mathbf{F}_{f}$, which captures the topological relationships between foreground instances, we further design a descriptor to exploit the extensive geometric information in the background point clouds.
Inspired by Scan Context~\cite{kim2022tro} encoding the maximum height of the raw point cloud into different bins of a polar BEV image, we leverage semantic information to replace the height data, constructing similar descriptors based on the background point cloud to generate rotation-invariant ring key descriptors.
This descriptor efficiently encodes the appearance characteristics of different background point clouds, offering a concise portrayal at a minimal computational cost.

Finally, \difffirst{we perform L2-normalization on $\mathbf{F}_{f}$ and $\mathbf{F}_{b}$ to enhance the robustness,} and concatenate them to form the LiDAR scan descriptor $\mathbf{F} \in \mathbb{R}^{D'}$. This comprehensive descriptor incorporates features from both the foreground semantic graph and the background, thus possessing stronger retrieval capabilities.
The LiDAR scan descriptor is utilized to retrieve the multiple similar candidate scans in the database.
The similarity between descriptors is computed using the Euclidean distance within the feature space.
In real applications, we use the FAISS library~\cite{Johnson2017tbd} to establish the descriptors database and perform fast parallel retrieving.

\vspace{-0.1cm}
\subsection{Geometric Verification}
\label{sec:geometric_vgerification}

Once loop candidates are obtained, geometric verification is performed on each candidate to determine whether a true loop closure has occurred. The core of geometric verification is to establish node correspondences and estimate an initial relative pose between the query and candidate scans. The similarity between them is then measured by assessing the degree of alignment, including the following four steps:

\textbf{Instance Nodes Matching.}
For the query scan with semantic graph $\mathcal{G}_q$ and the target candidate scan with semantic graph $\mathcal{G}_t$, we create an affiliation matrix $\mathbf{I} \in \RR^{N \times M}$ based on node descriptor similarity, where $N$ and $M$ are the number of nodes in the query and target graphs, respectively. Each element $\mathbf{I}_{ij}$ is the cost of assigning the $i$-th node in $\mathcal{G}_q$ to the $j$-th node in $\mathcal{G}_t$ as:
\begin{equation}
\mathbf{I}_{ij} = \begin{cases} 1-\frac{\mathbf{f}^{q}_i \cdot \mathbf{f}^{t}_j}{\Vert \mathbf{f}^{q}_i \Vert  \times \Vert \mathbf{f}^{t}_j \Vert } & \text{if} \;l_{vi}^{q}=l_{vj}^{t} \;\text{and}\; \mathbf{b}_i^{q}=\mathbf{b}_j^{t}, \\
                       10^8 & \text{otherwise}
         \end{cases}  
\end{equation} 
A high cost is assigned to reject a match if theirs node labels are inconsistent or bounding box sizes differ significantly, \difffirst{i.e., the difference in each dimension of the box beyond $d_b$.}

The matching results are then determined using the Hungarian algorithm~\cite{Kuhn1955nrl}, yielding the instance node matching correspondence set between $\mathcal{G}_q$ and $\mathcal{G}_t$ as:
\begin{equation}
    \mathcal{M}_q = \{\mathbf{c}^{q}_{1},\mathbf{c}^{q}_{2},...,\mathbf{c}^{q}_{o}\},
\end{equation} 
\begin{equation}
    \mathcal{M}_t = \{\mathbf{c}^{t}_{1},\mathbf{c}^{t}_{2},...,\mathbf{c}^{t}_{o}\},
\end{equation} 
where $\mathbf{c}=[x,y,z]^{\tr}$ is the position of node mentioned earlier, $o$ is the number of node matching pairs.

\textbf{Node Correspondences Pruning.}
Due to changes in viewpoint during revisit or errors in semantic segmentation, there are incorrect node correspondences.
Although RANdom SAmple Consensus (RANSAC)~\cite{fischler1981random} can exclude some outliers, using it with all node matches substantially increases the number of iterations, making it unsuitable for online applications.

To reduce the computational complexity, we propose an efficient outlier pruning scheme based on the local structure of nodes before applying RANSAC.
We posit that two nodes of a true positive correspondence should exhibit consistent local geometric structures.
For a node $\mathbf{v}$, its neighboring nodes within a certain range \difffirst{$d_l$} can be represented as $\{ \mathbf{v}_j \}_{j=1}^{K}$, where $K$ is the total number of neighbors.
$\mathbf{v}$ and any two of its neighbors can form a triangle, denoted as $\Delta_j$, and all local triangles of $\mathbf{v}$ is $\{ \Delta_j\}_{j=1}^{K(K-1)/2}$.
In an ideal scenario, corresponding local geometric triangles of matched nodes should be perfectly aligned. However, practical applications are subject to errors from semantic segmentation and clustering. Therefore, we relax this condition to prevent the elimination of correct correspondences.
For the matching of local triangles, we consider two triangles to be consistent when their corresponding sides are of equal length.
\difffirst{Once the number of successfully matched local triangles between two nodes exceeds $t_m$, they are considered as true node correspondences.}

\textbf{Initial Pose Estimation.}
Based on pruned correspondences, denoted as $\mathcal{M}_q^{'}$ and $\mathcal{M}_t^{'}$, we estimate a relative pose between the query and candidate scans based on RANSAC and SVD decomposition~\cite{Besl1992tpami}.
In each RANSAC iteration, we randomly select three matching pairs of $\mathcal{M}_q^{'}$ and $\mathcal{M}_t^{'}$ to solve the transformation equation based on SVD decomposition as:
\begin{equation}
\label{eq:align}
    \mathbf{R}_i, \mathbf{t}_i = \mathop{\text{min}}\limits_{\mathbf{R},\mathbf{t}} \sum\limits_{j=1}^{3} \Vert \mathbf{R} \cdot \mathbf{c}_{j}^{t} + \mathbf{t} - \mathbf{c}_{j}^{q}  \Vert_{2}^{2},
\end{equation} 
by which we obtain a transformation in each iteration and finally select the transformation with the greatest number of inliers from all iterations as a coarse pose estimation for loop candidate verification $\mathbf{T}_{\text{coarse}}=\{\mathbf{R}_c,\mathbf{t}_c\}$.

\textbf{Loop Candidate Verification.}
We verify the loop candidates by calculating scan similarity on two levels. First, we evaluate the semantic graph similarity between the query scan and candidate scan as follows:
\begin{equation}
    S_{\text{graph}} =  \text{exp}(-\frac{\sum\limits_{j=1}^{u} \Vert \mathbf{R}_c \cdot
    \mathbf{c}_{j}^{t}{'}  
    + \mathbf{t}_c
    - \mathbf{c}_{j}^{q}{'} 
    \Vert_2}{u}),
\end{equation} 
where $\mathbf{c}_{j}^{t}{'}$, $\mathbf{c}_{j}^{q}{'}$ are the inlier correspondences. $u$ is the number of inliers.
$S_{\text{graph}}$ measures the alignment of the two graphs.

Furthermore, we use $\mathbf{T}_{\text{coarse}}$ to align the background point clouds of the query and candidate scans, and then calculate the cosine similarity of their background descriptors $\mathbf{F}^{q'}_{b}$ and $\mathbf{F}^{t'}_{b}$.
This provides greater reliability compared to using graph similarity alone. If both $S_{\text{graph}}$ and the background similarity exceed thresholds \difffirst{$t_g$ and $t_b$}, the query scan and the candidate scan are considered a true loop closure. We then proceed to estimate their precise pose transformation as follows.

\subsection{6-DoF Pose Refinement}
\label{sec:pose}
The aforementioned $\mathbf{T}_{\text{coarse}}$, estimated at the sparse instance node level, may not be accurate and robust enough for closing the loop.
To enhance the accuracy of the relative pose estimation, we propose performing a dense points registration on all foreground instance points from the input semantic point cloud, using $\mathbf{T}_{\text{coarse}}$ as the initial transformation.

Benefiting from the already obtained instance node matches, finding the corresponding point matches between matched instances becomes easy and fast due to the significantly reduced search space.
In fact, $\mathbf{T}_{\text{coarse}}$ already initially align the query scan and candidate scan. Therefore, directly searching for the nearest neighbor points as point matches is sufficient to initialize and solve the Iterative Closest Point (ICP) as:
\begin{equation}
\label{eq:icp}
    \mathbf{T}_{\text{icp}} = \mathop{\text{min}}\limits_{\mathbf{T}} \sum\limits_{(\mathbf{p}_j^{q},\mathbf{p}_j^{t})\in \mathcal{C} } \Vert \mathbf{T} \mathbf{p}_j^{t} - \mathbf{p}_j^{q}  \Vert_{2}^{2},
\end{equation}
where $\mathcal{C}$ is the set of instance points nearest neighbor correspondences between query scan and loop scan.

The dense registration result $\mathbf{T}_{\text{icp}}$ encompasses the registration of all foreground instance points. 
However, background points, such as those from buildings and roads offering stable plane features, are also valuable for pose estimation. Therefore, to further optimize the pose accuracy from dense points registration, we utilize point-to-plane residuals to refine the final relative pose.
\begin{equation}
\label{eq:plane}
    \mathbf{T}_{\text{refine}} = \mathop{\text{min}}\limits_{\mathbf{T}} \sum\limits_{(\mathbf{p}_j^{q}{'},\mathbf{p}_j^{t}{'})\in \mathcal{C'} } \Vert \mathbf{n}_j^{\tr} (\mathbf{T} \mathbf{p}_j^{t}{'} - \mathbf{p}_j^{q}{'})  \Vert_{2}^{2},
\end{equation}
where $\mathcal{C'}$ is the set of point correspondences with consistent plane normal vector, denoted as $\mathbf{n}_j^{\tr} \in \RR^3$.

$\mathbf{T}_{\text{coarse}}$ is estimated via sparse node matching, while $\mathbf{T}_{\text{icp}}$ is determined by dense instance points registration. This \mbox{sparse-to-dense} registration mechanism enhances pose estimation accuracy in a \mbox{coarse-to-fine} manner.
Finally, the pose is further refined using point-to-plane constraints for background points. This semantically-guided coarse-fine-refine pose estimation process considers the alignment of both instance points and background points. 
By separately handling instance points and background points, each alignment stage begins with a favorable initial value, ensuring fast matching and convergence, as well as multi-step checked robust pose estimation.

\section{Experimental Evaluation}
\label{sec:exp}

\subsection{Experimental Setup}
\textbf{Dataset.} 
\difffirst{We evaluate our method on  KITTI~\cite{Geiger2013ijrr} and KITTI-360~\cite{liao2022tpami} datasets. 
Additionally, Ford Campus~\cite{Pandey2011ijrr} and Apollo~\cite{Lu2019cvpr} datasets are used to test its generalization ability.}

\textbf{Implementation Details.} 
\difffirst{The parameters used in our experiments are listed in~\tabref{tab:parameter}. 
Due to background similarity being easily affected by semantic segmentation quality, we set $t_b$ for different datasets ($t_b=0.5$ for KITTI, KITTI-360 and Apollo, $t_b=0.45$ for Ford Campus).  Other parameters remain unchanged.}
We use the semantic information from a semantic segmentation network~\cite{wang2024arxiv} on all datasets \difffirst{and we train different  models on the KITTI dataset to ensure that each test sequence remains unseen during the training phase}.
We mainly build semantic graphs from $static \; vehicle$, $pole$ and $trunk$ instances, and utilize background points from $building$, $fence$, $road$ and $vegetation$.
This is convenient for replacing or adding other semantic information.
\difffirst{To fairly compare all semantic-assisted loop closing methods, we evaluate them using the same labels from SegNet4D~\cite{wang2024arxiv}.}
All the experiments are conducted on a machine with an AMD 3960X @3.8GHz CPU and a NVIDIA RTX 3090 GPU. 
\difffirst{More implementation details can be found in our online supplementary~\footnote{https://github.com/nubot-nudt/SGLC/Supplementary.pdf}.}

\begin{table}[t]
    \color{black}
	 \caption{\difffirst{The parameters settings of SGLC.}}
    \renewcommand\arraystretch{0.9}
    \footnotesize
    \setlength\tabcolsep{2.6pt}
	\centering
	\begin{threeparttable}
	\begin{tabular}{ccc}
		\toprule
		Parameter & Value & Description   \\
        \midrule
        $d_{\text{max}}$ & 60m & edge threshold for semantic graph \\
        $d_i$ & 2m & interval for edge histogram descriptors \\
        $k$ & 30 & the dimension for eigenvectors \\
        $d_b$ & 2m &   difference threshold of bounding box dimension \\
        $d_l$ & 20m & the range for generating local triangle \\
        $t_m$ & 2 &  threshold for determining true node correspondences \\
        $t_g$ & 0.5 &  threshold for semantic graph similarity \\
        $t_b$ & 0.5 &  threshold for background similarity\\
        $D$ & 120 & the dimension for node descriptor\\
        $D'$ & 231 & the dimension for LiDAR scan descriptor\\

		\bottomrule
	\end{tabular}

	\end{threeparttable}
	\label{tab:parameter}
 \vspace{-0.4cm}
\end{table}

\begin{table*}[t]
	\caption{The performance comparison of F1 max scores and Extended Precision on the KITTI and KITTI-360 datasets.}
    \footnotesize
    \setlength\tabcolsep{4pt}
	\centering
	\begin{threeparttable}  
	\begin{tabular}{ll|ccccccc|cc|c}
		\toprule
         &  & \multicolumn{7}{c|}{KITTI}  & \multicolumn{2}{c|}{KITTI-360} & \difffirst{Time} \\ 
		\multicolumn{2}{c|}{Methods}  & \multicolumn{1}{c}{00} & \multicolumn{1}{c}{02} & \multicolumn{1}{c}{05} & \multicolumn{1}{c}{06} & \multicolumn{1}{c}{07} & \multicolumn{1}{c}{08} &  \multicolumn{1}{c|}{Mean}  & \multicolumn{1}{c}{0002} &  \multicolumn{1}{c|}{0009}  & \difffirst{(ms)}\\
		\midrule
         \multirow{6}{*}{\rotatebox{90}{DNN-based}} &PV & 0.779/0.641 & 0.727/0.691 & 0.541/0.536 & 0.852/0.767 & 0.631/0.591 & 0.037/0.500 & 0.595/0.621  & 0.349/0.515 & 0.330/0.510  & \difffirst{11.3}\\
       &\difffirst{SGPR$^\star$} & \difffirst{0.720/0.507} & \difffirst{0.823/0.531} & \difffirst{0.720/0.552} & \difffirst{0.680/0.524} & \difffirst{0.700/0.500} & \difffirst{0.683/0.506} & \difffirst{0.721/0.520} & \difffirst{0.862/0.506} & \difffirst{0.845/0.503} & \difffirst{8.8}\\
       & LCDNet & 0.970/0.847 & \textbf{0.966}/\textbf{0.917} & \textbf{0.969}/0.938 & 0.958/0.920 & 0.916/0.684 & \textbf{0.989}/0.908 & \underline{0.961}/0.869 & \textbf{0.998}/\textbf{0.993} & \underline{0.988}/0.734   & \difffirst{147.7}\\
        &OT & 0.873/0.800 & 0.810/0.725 & 0.837/0.772 & 0.876/0.809 &  0.625/0.505 & 0.667/0.510 & 0.781/0.687 &0.796/0.507  & 0.879/0.528  & \difffirst{22.8}\\
       & BEVPlace & 0.960/0.849 & 0.845/0.819 & 0.885/0.815 & 0.895/0.815 & 0.917/0.687 & 0.967/0.868 & 0.912/0.809 & 0.920/0.504  & 0.938/0.684  & \difffirst{33.9}\\
       & \difffirst{PADLoC$^\star$ } & \difffirst{\underline{0.983}/0.912} & \difffirst{0.920/0.880} & \difffirst{0.957/0.863} & \difffirst{0.981/0.944} & \difffirst{0.781/0.704} & \difffirst{0.910/0.718} & \difffirst{0.922/0.837} & \difffirst{0.948/0.596}  & \difffirst{0.969/0.797}  & \difffirst{221.5}\\
       \midrule
        \multirow{5}{*}{\rotatebox{90}{Handcrafted}}&SC  & 0.750/0.609 & 0.782/0.632 & 0.895/0.797 & 0.968/0.924 & 0.662/0.554 & 0.607/0.569 & 0.777/0.681 & 0.771/0.554 & 0.851/0.619  & \difffirst{13.9}\\
        &GOSMatch$^\star$ & \difffirst{0.916/0.535} & \difffirst{0.694/0.575} &\difffirst{0.785/0.611}& \difffirst{0.491/0.518}& \difffirst{\underline{0.947}/0.913}&0.908/0.571 & \difffirst{0.790/0.621} & 0.694/0.500 & 0.766/0.575 & \difffirst{\underline{6.4}}\\
        &SSC$^\star$ & \difffirst{0.955/0.865} & \difffirst{\underline{0.933}/0.768} & \difffirst{\underline{0.966}/0.956} & \difffirst{0.983/\underline{0.978}} & \difffirst{0.894/0.785} & \difffirst{0.950/\underline{0.940}} & \difffirst{0.947/0.882} & \underline{0.965}/0.793 & 0.966/\underline{0.873} & \difffirst{\textbf{4.6}}\\
        &BoW3D & 0.977/\underline{0.981} & 0.578/0.704 & 0.965/\textbf{0.969} & 0.985/\textbf{0.985} & 0.906/\underline{0.929} & 0.900/0.866 & 0.885/\underline{0.906}  & 0.210/0.560  &  0.682/0.761 & \difffirst{78.0}\\
        &\difffirst{CC} & \difffirst{0.977/0.964} & \difffirst{0.903/0.568} & \difffirst{0.958/0.901} & \difffirst{\underline{0.993}/0.959} & \difffirst{0.906/0.811} & \difffirst{0.823/0.575} & \difffirst{0.927/0.796} & \difffirst{0.717/0.560} & \difffirst{0.886/0.620} & \difffirst{8.4}\\
       \midrule
	  & Ours$^\star$ & \difffirst{\textbf{0.998}/\textbf{0.986}} & \difffirst{0.888/\underline{0.899}} & \difffirst{\textbf{0.969}/\underline{0.967}} & \difffirst{\textbf{0.995}/0.963} & \difffirst{\textbf{0.993}/\textbf{0.991}} & \underline{0.988}/\textbf{0.980} & \difffirst{\textbf{0.972}/\textbf{0.964}} & 0.932/\underline{0.934} & \textbf{0.994}/\textbf{0.978}  & \difffirst{10.2}\\
		\bottomrule
	\end{tabular}
	\begin{tablenotes}
	\footnotesize
    \item \text{ [ $F_1$ max / $EP$ ]}, the best results are highlighted in bold and the second best are underlined. $^\star$ denotes semantic-assisted method.
	\end{tablenotes}

	\end{threeparttable}
	\label{tab:performance_distance}
   \vspace{-0.35cm}
\end{table*}

\begin{table}[t]
	\caption{The performance evaluation of loop closure detection on the KITTI dataset using overlap-based loop pairs.}
    \footnotesize
    \renewcommand\arraystretch{0.9}
    \setlength\tabcolsep{9.5pt}
	\centering
	\begin{threeparttable}
	\begin{tabular}{cl|cccc}
		\toprule
		& Methods & AUC & F1max & \makecell[c]{Recall\\@1} &\makecell[c]{Recall\\@1\%}   \\
		\midrule
        \multirow{6}{*}{\rotatebox{90}{DNN-based}}& PV  &0.856 & 0.846 & 0.776 & 0.845 \\
       & \difffirst{SGPR$^\star$} & \difffirst{0.591} & \difffirst{0.575} & \difffirst{0.753} & \difffirst{0.980}  \\
       & LCDNet & 0.933 & 0.883 & 0.915 &0.974  \\
       & OT & 0.907 & 0.877 & 0.906 & 0.964 \\
       & BEVPlace & 0.926 & 0.889 &0.913 & 0.972 \\
       & \difffirst{PADLoC$^\star$} & \difffirst{\underline{0.934}} & \difffirst{\underline{0.903}} &\difffirst{0.930} & \difffirst{0.975} \\
       \midrule
      \multirow{5}{*}{\rotatebox{90}{Handcrafted}}& SC & 0.836 & 0.835 & 0.820 & 0.869 \\
       & GOSMatch$^\star$ &\difffirst{0.906} &\difffirst{ 0.829}& \difffirst{\underline{0.941}} & \difffirst{\textbf{0.997}} \\
       & SSC$^\star$ & \difffirst{0.924}& \difffirst{0.882} & \difffirst{0.900} & \difffirst{0.951}  \\
       & BoW3D & - & 0.893 & 0.807 &  - \\
       & \difffirst{CC} & \difffirst{0.873} & \difffirst{0.902} & \difffirst{0.865} &  \difffirst{0.868} \\
		\midrule
	  & Ours$^\star$ & \difffirst{\textbf{0.949}} & \difffirst{\textbf{0.931}} & \difffirst{\textbf{0.950}} & \difffirst{\underline{0.986}} \\
		\bottomrule
	\end{tabular}

	\end{threeparttable}
	\label{tab:performance_overlap}
\end{table}

\subsection{Loop Closure Detection}
\label{sec:loop_closure_detection}
We follow the experimental setups of Li~\etalcite{li2021iros} to evaluate LCD performance and regard LiDAR scan pairs as positive samples of loop closure when their Euclidean distance is less than 3\,m,  as negative samples if the distance exceeds 20\,m.
For the KITTI dataset, we perform performance comparisons on all sequences with loop closure.
For the KITTI-360 dataset, in line with~\cite{Cattaneo2022tro}, we focus on sequences 0002 and 0009 with the highest number of loop closures.
\difffirst{Besides, we also report the results on the Ford Campus dataset (sequence 00) and a subset of the Apollo Columbia Park MapData proposed by~\cite{chen2022ral} for generalization evaluation.}

\textbf{Metrics.}
Following~\cite{li2021iros,Cui2022ralO}, we use the maximum $F_1$ score and Extended Precision ($EP$) as evaluation metric.

\textbf{Baselines.}
We compare the results with SOTA baselines, including DNN-based methods: PointNetVLAD (PV)~\cite{Uy2018cvpr}, \difffirst{SGPR~\cite{Kong2020iros}}, LCDNet~\cite{Cattaneo2022tro}, OverlapTransformer (OT)~\cite{ma2022ral}, BEVPlace~\cite{Luo2023iccv}, \difffirst{PADLoC~\cite{Arce2023ral}}, as well as handcrafted-based methods: Scan Context (SC)~\cite{kim2022tro}, GOSMatch~\cite{zhuy2020iros}, SSC~\cite{li2021iros}, BoW3D~\cite{Cui2022ralO}, \difffirst{Contour Context (CC)~\cite{Jiang2023icra}}.

\textbf{Results.}
The results are shown in~\tabref{tab:performance_distance}. 
Our method outperforms SOTA on multiple sequences and achieves the best average F1max score and $EP$ for the KITTI dataset.
Specifically, for sequence 08 with many reverse loop closures, our approach still exhibits
superior performance, proving its good rotational invariance.
And on the KITTI360 dataset, our method remains competitive.
\difffirst{Additionally, we report the average runtime for all methods, including descriptor generation and loop closure determination. Our method performs as fast as handcrafted methods while achieving superior performance in most cases, demonstrating both efficiency and effectiveness.}

Furthermore, to investigate the loop closure detection performance at further distances, 
we followed~\cite{chen2021auro,ma2022ral} regarding two LiDAR scans as a loop closure when their overlap ratio is beyond 0.3, which indicates that the maximum possible distance between them is around 15\,m.
And we adopt the same experimental setup as theirs to evaluate our method on the KITTI 00 sequence using AUC, F1max, Recall@1, and Recall@1\% as metrics.
The results are shown in~\tabref{tab:performance_overlap}.
Due to BoW3D leaning more towards geometric verification, we are unable to generate its AUC and recall@1\% results from its open-source implementation.
From the results, our method significantly outperforms SOTA baselines in terms of overall metric, indicating its robustness.

\difffirst{To evaluate the generalization capability of SGLC, we conducted additional experiments on the Ford Campus and Apollo datasets, focusing on loop pairs with an overlap ratio greater than 0.3. We primarily compared semantic-assisted methods to assess their adaptability to scene semantic changes or label quality declines. Due to the page limitation, we show the semantic segmentation details in our online supplementary. 
As shown in the~\tabref{tab:generalization}, 
our method demonstrates the best performance on both the Ford Campus and Apollo datasets, highlighting its strong generalization capability.
When the quality of semantic segmentation declines, some instances may be partially segmented, but after clustering, they still form a single object, which does not affect the construction of our semantic graph.  
Once the semantic graph is built, the LiDAR scan descriptors retrieve multiple candidate scans, including the correct loop scan. Through our geometric verification via graph matching, the true loop closure is identified.
Besides, for short-term loop closing in SLAM, static vehicles can serve as valuable feature nodes to enhance loop detection.
}
\begin{table}[t]
    \color{black}    
	\caption{\difffirst{The generalization performance evaluation.}}
    \footnotesize
    \renewcommand\arraystretch{0.95}
    \setlength\tabcolsep{2.9pt}
	\centering
	\begin{threeparttable}
	\begin{tabular}{l|cccc|cccc}
		\toprule
        & \multicolumn{4}{c|}{Ford Campus} &  \multicolumn{4}{c}{Apollo}\\
		 Methods & AUC & F1max & \makecell[c]{Recall\\@1} &\makecell[c]{Recall\\@1\%}  & AUC & F1max & \makecell[c]{Recall\\@1} &\makecell[c]{Recall\\@1\%}  \\
		\midrule
        SGPR& 0.412 & 0.439 & 0.467 & 0.951 & 0.640& 0.451 & 0.626 & 0.936 \\
        GOSMatch&0.752 & 0.632 & 0.820 & 0.954 & 0.677& 0.568 & 0.739 & 0.934 \\
        SSC & 0.924 & 0.865 & \underline{0.915} & \underline{0.964} & 0.937 & \underline{0.916} & 0.917 & 0.943 \\
        PADLoC & 0.938 & 0.873 & 0.910 &0.963 & 0.721 & 0.609 & 0.735 & 0.910  \\
        \midrule
       Ours$^\dag$ & \underline{0.955} & \textbf{0.908} &\textbf{0.920} & \textbf{0.970 }& \underline{0.956} & 0.910  & \underline{0.933} & \underline{0.951}\\
       Ours  &  \textbf{0.959}& \underline{0.897} & 0.896 & 0.960 & \textbf{0.957} & \textbf{0.919}  & \textbf{0.956} & \textbf{0.974} \\
		\bottomrule
	\end{tabular}
	\begin{tablenotes}
	\footnotesize
	\item $^\dag$ denotes build semantic graph without vehicle node. The best results are highlighted in bold and the second best are underlined.
    
	\end{tablenotes}

	\end{threeparttable}
	\label{tab:generalization}
 \vspace{-0.2cm}
\end{table}

\begin{table*}[t]
	\caption{The comparison of loop pose estimation errors on the KITTI dataset}
    \footnotesize
    \setlength\tabcolsep{4.5pt}
	\centering
	\begin{threeparttable}
	\begin{tabular}{l|ccc|ccc|ccc|ccc|c}
		\toprule
        & \multicolumn{6}{c|}{$Distance-based \quad closed \quad loop$}
        & \multicolumn{6}{c|}{$Overlap-based \quad closed \quad loop$}
        & \\
		& \multicolumn{3}{c|}{Seq.00}  & \multicolumn{3}{c|}{Seq.08} 
         & \multicolumn{3}{c|}{Seq.00}  & \multicolumn{3}{c|}{Seq.08}
        & \\
       & RR(\%) & RTE(m)  & \difffirst{RRE($^\circ$)} & RR(\%) & RTE(m)  & \difffirst{RRE($^\circ$)}
       & RR(\%) & RTE(m)  & \difffirst{RRE($^\circ$)} & RR(\%) & RTE(m)  & \difffirst{RRE($^\circ$)}
       & Time(ms)\\
       \midrule
      BoW3D &  \difffirst{95.61} & \underline{0.06} &  \difffirst{0.81} &  \difffirst{77.29} & \underline{0.09} &  \difffirst{2.06}
      &  \difffirst{57.69} & \underline{0.07} & \difffirst{0.92} & \difffirst{46.10} & \underline{0.10} & \difffirst{1.95} & \difffirst{\underline{40.0}}\\
      LCDNet(fast) & \difffirst{97.44}& \difffirst{0.52} & \difffirst{0.60} & \difffirst{78.28} & \difffirst{0.99} & \difffirst{1.29}
      & \difffirst{66.70} & 0.56 & \difffirst{1.02} & \difffirst{47.20} & 0.88& \difffirst{1.29} & \difffirst{81.7}\\
      LCDNet & \textbf{100} & \difffirst{0.13} & \difffirst{\underline{0.44}} & \textbf{100} & \difffirst{0.20}  & \difffirst{\underline{0.57}}
      & \difffirst{\underline{93.51}} & 0.21 & \difffirst{\underline{0.81}}& \difffirst{\underline{95.39}} & 0.31  & \difffirst{\underline{0.94}}& \difffirst{429.1}\\
      Ours & \textbf{100} & \textbf{0.04} & \difffirst{\textbf{0.21}} & \textbf{100} & \difffirst{\textbf{0.08}} & \difffirst{\textbf{0.41}} 
      & \difffirst{\textbf{99.82}} &\textbf{0.04}& \difffirst{\textbf{0.24}}& \difffirst{\textbf{99.57}} & \difffirst{\textbf{0.08}} & \difffirst{\textbf{0.46}} & \difffirst{\textbf{20.8}}\\
      \midrule
	\end{tabular}
	\begin{tablenotes}
	\footnotesize
    \item  The best results are highlighted in bold and the second best are underlined.
	\end{tablenotes}

	\end{threeparttable}
	\label{tab:loop_kitti}
    \vspace{-0.4cm}
\end{table*}

\begin{table}[t]
	\caption{Loop pose estimation results on the KITTI-360 dataset. }
    \footnotesize
    \setlength\tabcolsep{4pt}
	\centering
	\begin{threeparttable}
	\begin{tabular}{l|ccc|ccc}
		\toprule
		& \multicolumn{3}{c|}{Seq.0002}  & \multicolumn{3}{c}{Seq.0009} \\
       & RR(\%) & RTE(m)  & \difffirst{RRE($^\circ$)} & RR(\%) & RTE(m)  & \difffirst{RRE($^\circ$)}  \\
       \midrule
      BoW3D & \difffirst{67.73} & \difffirst{\textbf{0.20}}  & \difffirst{1.99}& \difffirst{87.32} & \underline{0.13} & \difffirst{1.31} \\
      LCDNet(fast) & \difffirst{82.92}& 0.84 & \difffirst{1.17} & \difffirst{98.81} &\difffirst{ 0.42} & \difffirst{1.07}   \\
      LCDNet & \difffirst{\textbf{98.56}} & 0.28&\difffirst{ \underline{1.03}} & \textbf{100} & \difffirst{0.19}  & \difffirst{\underline{0.74}} \\
      Ours &\difffirst{\underline{95.01}} &\difffirst{\underline{0.21}}  &\difffirst{\textbf{0.89}}  & \difffirst{\underline{99.40}} & \textbf{0.11} & \difffirst{\textbf{0.51}} \\
      \bottomrule
	\end{tabular}

	\end{threeparttable}
	\label{tab:loop_correction_kitti360}
 \vspace{-0.2cm}
\end{table}

\vspace{-0.2cm}
\subsection{Loop Pose Estimation}
To validate the loop pose estimation performance, we follow Cattaneo~\etalcite{Cattaneo2022tro} and evaluate our approach on the KITTI sequences 00 and 08 and KITTI-360 sequences 0002 and 0009.
To keep consistent with them, we choose the loop closure samples when the distance is within 4\,m.
Additionally, we also select the more challenging loop pairs with overlap rate beyond 0.3 for testing.

\textbf{Metric.}
(i) Registration Recall (RR), represents the percentage of successful alignment.
(ii) Relative Translation Error (RTE), the Euclidean distance between the ground truth and estimated poses.
\difffirst{(iii) Relative Rotation Error (RRE), the rotation angle difference between the ground truth and estimated poses.}
Two scans are regarded as a successful registration when the RTE and RRE are below 2\,m and 5$^\circ$, respectively.

\textbf{Baselines.}
The baseline methods for comparison mainly include 6-DoF pose estimation methods
BoW3D~\cite{Cui2022ralO} and LCDNet~\cite{Cattaneo2022tro}.
LCDNet is available in two version: LCDNet (fast) utilizes an unbalanced optimal transport (UOT)-based head to estimate relative pose while LCDNet replaces the head by a RANSAC estimator.

\begin{figure}[t]
	\centering
	\includegraphics[width=1\linewidth]{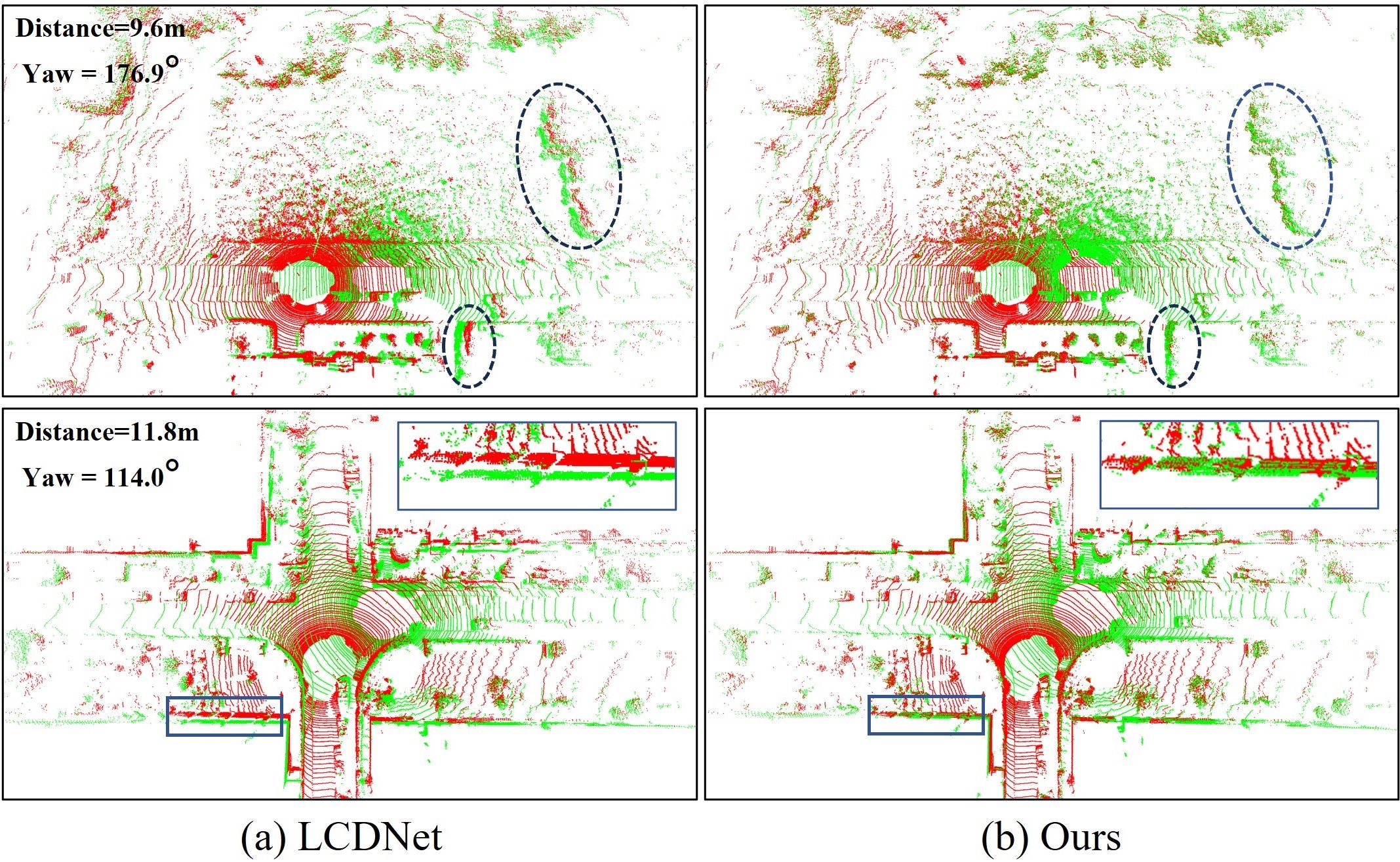}
	\caption{The qualitative comparison of loop pose estimation on the KITTI dataset using overlap-based loop pairs. Dashed ellipses are directly annotated on the registration results, while solid boxes indicate local magnification. \difffirst{The top-left corner displays the ground truth distance and yaw angle difference of the loop pair.}
	}
	\label{fig:qualitative}
	\vspace{-0.5cm}
\end{figure}

\textbf{Results.} 
\difffirst{As shown on the~\tabref{tab:loop_kitti}, Our method achieves the best loop pose estimation performance on both distance-based loop pairs and more challenging overlap-based loop pairs.
Initially, both LCDNet and our method achieve a 100\% RR for distance-based loop pairs.
However, for overlap-based loop pairs, LCDNet's performance declined, while our method still maintains superior registration capabilities, demonstrating its robustness for viewpoint changes.
In terms of alignment accuracy, our method also significantly outperforms other baselines.}
We also present the qualitative results in~\figref{fig:qualitative}. 
As can be seen,  our method achieves better alignment even in low overlap loop closure, attributing to the semantically-guided coarse-fine-refine point cloud alignment process.
\tabref{tab:loop_correction_kitti360} shows the results on the KITTI-360 dataset with distance-based close loop. 
\difffirst{It can be observed that LCDNet exhibits the best generalization ability, but our method remains competitive and  has the best alignment accuracy, considering both RTE and RRE.}
We also report the average running time of pose estimation on our machine.
From the comparison, our method boasts the fastest running speed.
\difffirst{The total runtime of our system, incorporating semantic segmentation, loop closure detection and poses estimation, is 98.1~ms, less than the data acquisition time of 100~ms for a typical rotational LiDAR sensor, thereby rendering it suitable for real-time SLAM systems.}

\vspace{-0.2cm}
\subsection{Performance on SLAM System}
\begin{figure}[t]
	\centering
	\includegraphics[width=1\linewidth]{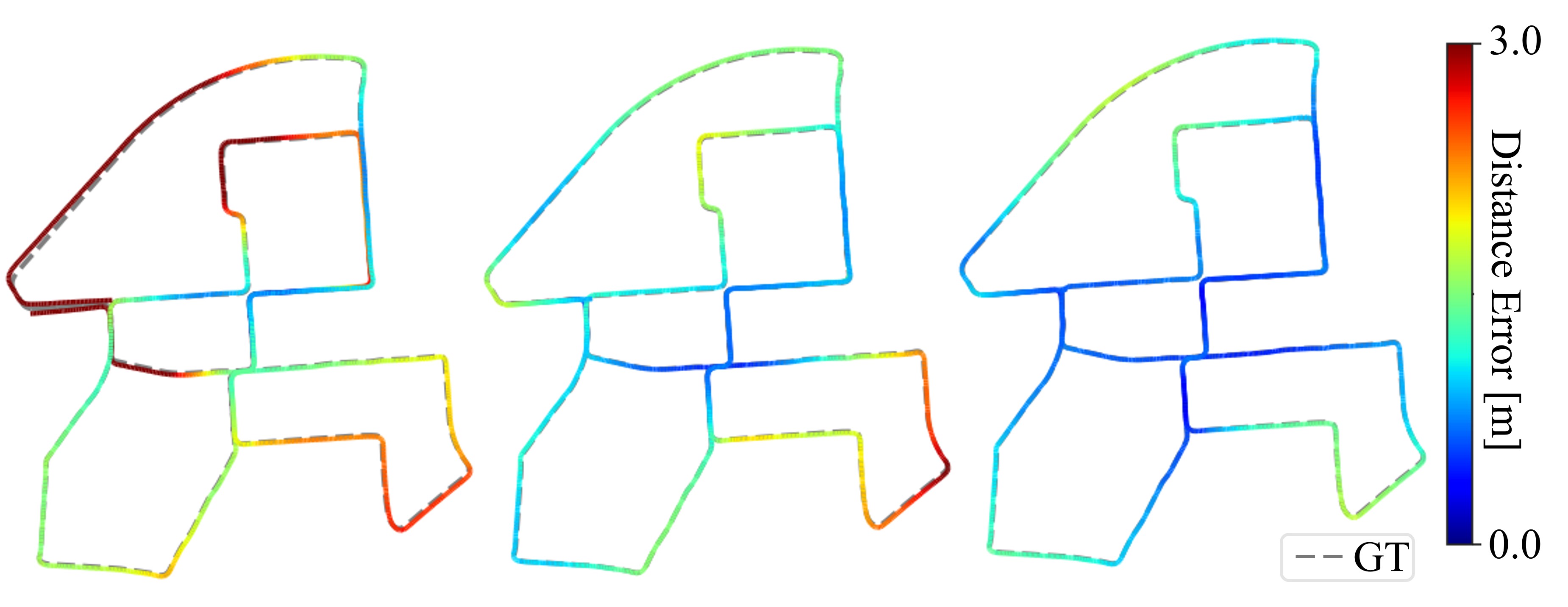}
	\caption{The trajectory of A-LOAM odometry (left) compared to with integrating BoW3D loop closing method (middle) and integrating our approach (right) on the KITTI 00 sequence.  
	}
	\label{fig:salm}
	\vspace{-0.25cm}
\end{figure}

We integrate SGLC into A-LOAM odometry to eliminate cumulative errors for evaluating the accuracy of SLAM trajectories, and compare with baseline integrated with BoW3D.
We utilize a Incremental Smoothing and Mapping (iSAM2)~\cite{Kaess2012ijrr} based pose-graph optimization (PGO)\footnote{https://github.com/gisbi-kim/SC-A-LOAM} framework to build a factor graph for managing the the keyframe poses.
If a loop closure is detected, we will add this loop closure constraint into the factor graph to eliminate errors and ensure global trajectories consistency.
As shown in~\figref{fig:salm},  we present the trajectory error analysis between keyframe and ground truth poses.
From the results, our method significantly enhances the SLAM system by reducing the cumulative odometry error, and shows superior performance compared to the baseline that incorporates BoW3D. This improvement is attributed to the excellent loop closing capability of SGLC.

\vspace{-0.1cm}
\subsection{Ablation Study}
In this section, we conduct a series of ablation studies \difffirst{on KITTI sequence 00 and 08 and report the mean metrics to evaluate the effectiveness of designed components, and analyze their runtime.
The results are presented in~\tabref{tab:ablation}.
$[$A$]$ is our baseline by removing those components listed in~\tabref{tab:ablation}.
Comparing  $[$A$]$ and $[$B$]$, it is demonstrated that incorporating the global properties of instance nodes can enhance the performance of loop closing.
As anticipated, it increases each node's distinctiveness, which is beneficial for finding the correct node correspondences.
For $[$C$]$, we can see the effectiveness of outlier pruning in improving registration accuracy with the same number of RANSAC iterations as $[$B$]$.
In experiment $[$D$]$, $[$E$]$ and $[$F$]$, dense points registration and point-to-plane alignment are applied mainly for loop scan determined after geometric verification, so they do not affect the results of loop closure detection.
The results clearly show that both dense points registration and point-to-plane alignment can improve registration performance, and we can get optimal results only when they operate in conjunction.
The comparison of $[$F$]$, $[$G$]$, $[$H$]$ also indicates that even direct point-to-point ICP or point-to-plane alignment on the raw scan cannot get optimal results and significantly increase computational costs on direct point-to-point ICP, which demonstrates that our registration process effectively utilizes foreground and background.
Besides, each component in our framework exhibits high execution efficiency.}


\begin{table}[t]
	\footnotesize
        \renewcommand\arraystretch{0.8}
        \setlength{\tabcolsep}{2.5pt}
	\caption{Ablation studies. “Node Glo.” denotes the global properties of instance nodes. “Out. Pru.” denotes outlier pruning for node matches. “Den. ICP” denotes dense points registration, i.e., $\mathbf{T}_{\text{icp}}$. “P-P Ali.” denotes point-to-plane alignment, i.e., $\mathbf{T}_{\text{refine}}$. }
	\centering
	\begin{threeparttable}
		\begin{tabular}{l|p{0.7cm}<{\centering}p{0.6cm}<{\centering}p{0.6cm}<{\centering}p{0.6cm}<{\centering}|p{0.8cm}<{\centering}p{0.8cm}<{\centering}p{0.9cm}<{\centering}p{0.9cm}<{\centering}|p{0.7cm}<{\centering}}
		\toprule
			&\makecell[c]{Node\\Glo.}  &   \makecell[c]{Out.\\Pru.}  & \makecell[c]{Den.\\ICP} & \makecell[c]{P-P\\Ali.} &  F1max &  RR(\%) & RTE(m) &  \difffirst{RRE($^\circ$)} &\makecell[c]{Time\\(ms)}  \\
			\midrule
            \rowcolor{LightGrey}
			$[$A$]$&    &    &  &  &   \difffirst{0.989}  &  \difffirst{97.30}  &  \difffirst{0.21} &  \difffirst{1.23} &   \difffirst{9.1}\\
            $[$B$]$& \Checkmark  &  &   & &   \difffirst{0.993}  &  \difffirst{97.76}  & \difffirst{0.20} &  \difffirst{1.17} &  \difffirst{+0.5}\\
            $[$C$]$&   \Checkmark &  \Checkmark   & & &   \difffirst{\textbf{0.994}}  &  \difffirst{98.29}  &  \difffirst{0.18} & \difffirst{0.99} &   \difffirst{+1.2}\\
            $[$D$]$&  \Checkmark &   \Checkmark & \Checkmark &   &  -  &  \difffirst{99.98}  &  \difffirst{0.07} & \difffirst{0.38} &  \difffirst{+11.8}\\
            $[$E$]$&  \Checkmark &  \Checkmark  &  & \Checkmark  &  -  &  \difffirst{99.71} &  \difffirst{\textbf{0.06}} & \difffirst{0.34} &  \difffirst{+8.5}\\
            $[$F$]$&  \Checkmark &  \Checkmark  & \Checkmark & \Checkmark  &  -  &  \difffirst{\textbf{100}}  &  \difffirst{\textbf{0.06}} & \difffirst{\textbf{0.31}} & \difffirst{ +20.2}\\
            \midrule
            \difffirst{$[$G$]$}&   \difffirst{\Checkmark} &   \difffirst{\Checkmark}  &  \difffirst{$\bullet$} &   &   \difffirst{-}  &  \difffirst{\textbf{100}}  &  \difffirst{\textbf{0.06}} &  \difffirst{0.33} &  \difffirst{+86.1}\\
             \difffirst{$[$H$]$}&   \difffirst{\Checkmark} &   \difffirst{\Checkmark}  &  &  \difffirst{$\bullet$}  &   \difffirst{-}  &  \difffirst{99.65}&  \difffirst{\textbf{0.06}} & \difffirst{0.34} &  \difffirst{+9.4}\\
			\bottomrule
		\end{tabular}
	\end{threeparttable}
    \begin{tablenotes}
	\footnotesize
	\item   +~denotes the time increment compared to $[$A$]$.  \difffirst{$\bullet$~denotes direct point-to-point ICP or point-to-plane alignment for the raw scan.}
	\end{tablenotes}
	\label{tab:ablation}
 \vspace{-0.4cm}
\end{table}

\section{Conclusion}
\label{sec:conclusion}
This paper presents a semantic graph-guided  approach for LiDAR-based loop closing.
It builds a semantic graph of the foreground instances for descriptor generation and guides subsequent geometric verification and pose estimation.
The designed LiDAR scan descriptor leverages both the semantic graph's topological properties and the appearance characteristics of the background points, thereby enhancing its robustness in LCD.
For pose estimation, we proposed a \mbox{coarse-fine-refine} registration scheme that considers the alignment of both instance and background points, offering high efficiency and accuracy.
The method supports real-time online operation and is convenient for integration into SLAM systems.
The experimental results on multiple datasets demonstrate its superiority.

\vspace{-0.1cm}
\bibliographystyle{unsrt}
\bibliography{new}
\end{document}